\documentclass[11pt]{article}

\usepackage{algorithm}
\usepackage[noend]{algpseudocode}
\usepackage{multirow}
\usepackage{color}
\usepackage{amsmath}
\usepackage{amsfonts}
\usepackage{graphicx}
\usepackage{booktabs}
\usepackage{url}
\usepackage[tight,footnotesize]{subfigure}
\usepackage[a4paper, total={6in, 8in}]{geometry}

\title{Genetic-based Constraint Programming for Resource Constrained Job Scheduling}

\author{Su Nguyen$^1$ \and Dhananjay Thiruvady$^2$ \and Yuan Sun$^3$ \and Mengjie Zhang$^4$}

\date{
	$^1$School of Accounting, Information Systems and Supply Chain, RMIT University \\ \texttt{su.nguyen@rmit.edu.au}\\[2ex]
	$^2$School of Information Technology, Deakin University \\ \texttt{dhananjay.thiruvady@deakin.edu.au}\\[2ex]%
    $^3$La Trobe Business School, La Trobe University \\ \texttt{yuan.sun@latrobe.edu.au}\\[2ex]%
    $^4$Centre for Data Science and Artificial Intelligence, Victoria University of Wellington \\ \texttt{mengjie.zhang@ecs.vuw.ac.nz}\\[2ex]%
}

\begin{document}

\maketitle

\begin{abstract}
Resource constrained job scheduling is a hard combinatorial optimisation problem that originates in the mining industry. Off-the-shelf solvers cannot solve this problem satisfactorily in reasonable timeframes, while other solution methods such as many evolutionary computation methods and matheuristics cannot guarantee optimality and require low-level customisation and specialised heuristics to be effective.  This paper addresses this gap by proposing a genetic programming algorithm to discover efficient search strategies of constraint programming for resource-constrained job scheduling. In the proposed algorithm, evolved programs represent variable selectors to be used in the search process of constraint programming, and their fitness is determined by the quality of solutions obtained for training instances. The novelties of this algorithm are (1) a new representation of variable selectors, (2) a new fitness evaluation scheme, and (3) a pre-selection mechanism. Tests with a large set of random and benchmark instances, the evolved variable selectors can significantly improve the efficiency of constraining programming. Compared to highly customised metaheuristics and hybrid algorithms, evolved variable selectors can help constraint programming identify quality solutions faster and proving optimality is possible if sufficiently large run-times are allowed. The evolved variable selectors are especially helpful when solving instances with large numbers of machines.
\end{abstract}


\section{Introduction}

In the field of scheduling, a number of problems can be formulated as job scheduling problems. Among these, the variant -- resource-constrained job scheduling (RCJS) with shared resources -- is a typical example of a real-world application. The problem was originally motivated in the mining supply chain and aims to optimise the throughput of ore from mines to ports. The specific components of the RCJS problem and their relationship to mining are as follows. A single batch of ore that needs to be transported is a job. The modes of transport are typically trains or trucks, which are limited in number. Therefore, batches of ore that need to be transported concurrently must share the trains and trucks, and in RCJS, this is modelled by the resource limit on jobs running in parallel. Ore batches may also need to arrive in a particular order at ports, which is effectively modelled by imposing precedences between jobs. A major concern is the untimely arrival of ore batches at ports, which results in demurrage costs. It is nearly impossible to ensure that all batches of ore arrive in a timely fashion, but limiting untimely arrivals can lead to large improvements in the mining supply chain. This aspect can be modelled as an objective that minimises the total weighted tardiness (TWT) across all jobs. 

Due to the complexity of RCJS, exact approaches on their own have not been effective. Moreover, the complicating constraints mean that metaheuristics on their own struggle to find good feasible regions. Hence, a number hybrid methods and in particular matheuristics have proved most effective \cite{singhernst10,ernstsingh12,Thiruvady2014,Thiruvady2016,nguyen_cor_2019,blum2019,Thiruvady2020ms,Nguyen2022}. The original studies on this problem investigate Lagrangian relaxation and column generation based matheuristics, where integer programming decompositions are combined with simulated annealing, particle swarm optimisation and ant colony optimisation \cite{singhernst10,ernstsingh12,Thiruvady2014}. All these studies demonstrate that good feasible solutions can be found in reasonable time frames, and moreover good lower bounds to the problem can obtained. However, solving large problems was still a challenge, and hence parallel implementations were attempted and showed promise \cite{Thiruvady2016,Cohen2017,nguyen_cor_2019}. A recent study by \cite{blum2019} applies a biased random key genetic algorithm, and shows that excellent results can be found in short time frames. To the authors' knowledge, the most recent study by \cite{Thiruvady2020ms} proposes a new matheuristic, merge search, which proves to be very effective on small to medium sized problems.

Previous studies have also shown that constraint programming (CP) can be effective on variants of RCJS \cite{Thiruvady2012,Cohen2017}. \cite{Thiruvady2012} showed that CP on its own requires substantial resources, and hence, they propose a hybrid of CP with ant colony optimisation and beam search. While this approach proves to be effective, it still suffered from large run-time requirements. The study by \cite{Cohen2017} further aimed to improve efficiency by considering parallel implementations for RCJS, though there was still significant room for improvement. 

A recent line of research has explored the interface between optimisation and machine learning, and among the approaches that are gaining popularity, automated heuristic design (AHD) is of substantial interest~\cite{branke_automated_2016,pillay2018,Nguyen2017,zhang2020FS}. 
AHD is a hyperheuristic method that departs from traditional optimisation approaches \cite{burke_evolving_2006,vanLon2018,Kieffer20,HaoHyper20}, where the aim is to explore the space of heuristics rather than the traditional aim of exploring a solution search space. Several machine learning techniques have been effectively applied for learning or evolving heuristics, including logistic regression \cite{ingimundardottir_supervised_2011}, decision trees \cite{olafsson_learning_2010} and artificial neural networks \cite{mouelhi-chibani_training_2010,bello17}. An alternative is genetic programming (GP), a popular method for AHD. The method maintains a population of programs (depending on fitnesses), which are created and evolved using the crossover and mutation operators across a number of generations. Typically, a meta-algorithm is used for fitness evaluations, where the meta-algorithm is a solution construction procedure or meta-heuristic template, and these are used to solve a number of simulated scenarios or training instances \cite{Nguyen2017}. The GP approach has two main strengths, which are search mechanisms and flexible representations, allowing it to deal very effectively with potentially many decisions and technical requirements inherent in optimisation problems. GP has been widely used in scheduling and combinatorial optimisation, including resource allocation \cite{CHAND2018146}, dynamic pickup and delivery \cite{vanLon2018}, production scheduling \cite{pickardt_evolutionary_2013,zhang2019two}, and web composition \cite{Silva16}. The strength of the method is evidenced in these studies, which have shown for a number of applications, that the evolved GP-based programs or heuristics it finds, often outperform manually designed heuristics identified by problem domain experts.

The combination between GP and CP has recently been investigated by \cite{Nguyen21GCP}. In their study, the proposed GP algorithm is used to evolve variable selectors in CP for job shop scheduling, and the fitness of a variable selector is the number of branching steps in CP needed to reach optimal solutions. The results show that the evolved variable selectors can help CP quickly identify good solutions and increase the chance of finding optimal solutions. Unlike most applications of GP for combinatorial optimisation problems, no meta-algorithm is needed for solution construction (or improvement) as their algorithm can directly utilise the problem formulation and CP to construct complete solutions and improve their quality. 


In this paper, we introduce the first investigation into the combination of GP and CP for the RCJS problem. The aim of this study is to examine the extent to which GP can improve the performance of CP in solving the RCJS problem. We build upon the approach developed in \cite{Nguyen21GCP} and propose a novel GP algorithm that evolves the CP variable selectors for solving the RCJS problem. Our algorithm introduces three key advancements: (1) a novel representation of variable selectors for RCJS, (2) a new fitness evaluation scheme, and (3) a pre-selection mechanism. These advancements are designed to address the challenges of efficiently evolving variable selectors for RCJS using GP, which cannot be adequately addressed by the original algorithm developed in \cite{Nguyen21GCP}.  

The rest of this paper is organised as follows. Section~\ref{sec:background} describes the RCJS problem and discusses related work. Section~\ref{sec:gcp} presents the proposed GP algorithm and its key components. Datasets and parameter settings are presented in Section~\ref{sec:exp}. Section~\ref{sec:results} presents the experiment settings and the results of the proposed algorithm compared to a metaheuristic and a hybrid method. Finally, conclusions and future work are presented in Section~\ref{sec:conclude}.
\section{Background}
\label{sec:background}
\subsection{Problem formulation}

The RCJS can be formally defined as follows. There are a number of machines $\mathcal M = \{m_1,\ldots,m_l\}$, jobs $\mathcal J$ $= \{j_1,\ldots,j_n\}$ and time horizon $\mathcal T$, given. A job $i$ has with it the following data associated:
\begin{itemize}
	\item $r_i$: release time
	\item $p_i$: processing time
	\item $d_i$: desired due time
	\item $w_i$: the weight of the job
    \item $g_i$: amount of resource needed
	\item $m_i$: the machine that the job must be scheduled on
\end{itemize}

Moreover, a set of precedences $\mathcal P$ are given, where between two jobs $i$ and $j$ on the same machine, a precedence relation may exist ($i \rightarrow j$ implies that job $i$ must complete before job $j$ commences). Moreover, resource limits are imposed with the renewable resource $R$, where all jobs executing at any time point concurrently, must not cumulatively use more than the limit $R$. The objective is to minimize total weighted tardiness
\begin{equation}
	T({\bf \pi}) = \sum_{i=1}^{n} w_{\pi_i} \times T(\pi_i)
\end{equation}
\noindent where $\pi$ is a permutation of the jobs and $T(\pi_i)$ is tardiness of job $i$. 


The RCJS problem can be formulated as a constraint program as follows. We define the start time and end time variables for all jobs, $s_j$ and $e_j$ for job $j$, respectively.  
 \begin{align}
     \forall j \in {\mathcal J}: & \quad &  s_{j} \geq r_j \label{cp:one}\\
     \forall j \in {\mathcal J}: & \quad &  e_{j} = s_{j} + p_{j} \label{cp:two}\\
     \forall i,j \in {\mathcal J}: & \quad & m_i = m_j \Rightarrow s_i > e_j \vee s_j > e_i  \label{cp:three} \\
     \forall (i,j) \in {\mathcal P}: & & \quad s_{j} > e_i \label{cp:four}
\end{align}

Constraint~\ref{cp:one} ensures a job starts after it is released. Constraint~\ref{cp:two} links the start and end time variables for all jobs. Constraint~\ref{cp:three} are disjunctive constraints, and ensure that jobs on the same machine cannot execute at the same time. The precedences between two tasks on the same machine are imposed by Constraint~\ref{cp:four}.

Finally, to ensure the resource constraints are satisfied, we make use of the solver's in-built high-level constraint, {\it cumulatives($s$,$p$,$e$,$r$,$R$)}. This constraint takes as input the start ($s$) time variables, end time ($e$) variables, and data given by the processing times ($p$), resource requirements ($r$) and resource limit ($R$). The start and end time variables are determined by considering $p$, $r$ and $R$, so that if two or more tasks overlap in their execution at any time point (irrespective of their machine), they must cumulatively use at most resource availability $R$. 

\subsection{Optimisation algorithms for RCJS}
A closely related problem to the RCJS that has been studied extensively is project scheduling \cite{bruker99}. Both problems consist of several similarities, including jobs with known processing times, shared resources, and precedence relations. There are also important differences, such as the objectives. Therefore, approaches to solving project scheduling can also be considered for RCJS, and numerous such approaches have been applied ~\cite{ballestin08,bruker99,demeulemeester02,schwindt03,Thiruvady2014ps,Brent14,Thiruvady2019}. Exact approaches on their own are often infeasible in terms of time, and hence most of these methods focus on metaheuristics, hybrid methods, matheuristics, decompositions, and parallel implementations. \cite{demeulemeester02} investigates exact and incomplete approaches and finds that genetic algorithms and simulated annealing are most effective on variant of the problem. \cite{schwindt03} investigate project scheduling with time windows and devise exact and heuristic approaches to solve it. \cite{ballestin08} investigate a project scheduling variant that is closely related to RCJS, where the objective is to minimise the cumulative deviation of the completion times of all tasks. They show that a local search approach is the most effective on this problem. The net present value objective has also been considered by recent studies~\cite{Thiruvady2014ps,Brent14,Thiruvady2019}, which show that meta-heuristics and parallel implementations can be used to achieve high levels of efficiency.

Of particular interest in RCJS, is the TWT objective.  This objective has indeed been considered in several similar problems~\cite{singhweiskircher10,singhweiskircher11}. The studies by \cite{singhweiskircher10} and \cite{singhweiskircher11} use agent-based modelling with relatively little information sharing. These studies also assume decentralised data. Nonetheless, these studies demonstrate the complexity of RCJS type problems, especially considering the TWT objective. 

Recently, a variant of RCJS with uncertainty has been investigated \cite{THIRUVADY2022sacs, thiruvady2022adaptive}. \cite{THIRUVADY2022sacs} show that a population-based ant colony optimisation with surrogate models can provide excellent solutions to the problem. Furthermore, \cite{thiruvady2022adaptive} propose an adaptive simulated annealing approach, which outperforms population-based ant colony optimisation and also proves to be effective on the original RCJS problem.  

We see that several studies have been carried out on RCJS and related problems, but there is still considerable scope for improvement, particularly in terms of solution approaches. The most successful methods to date have been metaheuristics, mixed integer programming, constraint programming and matheuristics. However, there has been little progress in integrating the benefits of machine learning and optimisation for tackling such complex problems. This study aims to make progress in this direction.

\subsection{Machine learning for combinatorial optimisation}
Machine learning (ML) has emerged as a powerful tool to tackle combinatorial optimisation problems. Integrating ML into combinatorial optimisation focusses on enhancing traditional algorithms with advanced decision-making capabilities \cite{bengio2018machine}. This synergy is particularly evident in how ML can approximate complex computations or develop new policies for decision making, as highlighted in research on discrete optimisation problems like integer-constrained optimisation. The methodologies employed in the literature vary from supervised learning, where ML models mimic expert decisions, to reinforcement learning, which involves learning and optimising decision functions from scratch.

The integration of ML into combinatorial optimisation can be categorised into three distinct approaches \cite{bengio2018machine}. The first, \textit{end-to-end learning}, involves using ML models to directly output solutions from given problem instances, with the ML model acting independently to solve discrete optimisation problems \cite{bello2016neural}. The second approach, \textit{learning to configure algorithms}, uses ML to enhance traditional combinatorial algorithms. Here, ML aids in configuring algorithm parameters or providing additional decision-making information, thereby optimising the algorithm's performance \cite{bonami2018learning}. The third approach, \textit{machine learning alongside optimisation algorithms}, represents a more collaborative interplay between ML and combinatorial optimisation.

GP has been applied to many combinatorial optimisation problems from production scheduling \cite{nguyen_computational_2013} to dynamic pickup and delivery \cite{vanLon2017}. The main goal of GP in these studies is to automatically discover good heuristics for either constructing solutions (similar to ``end-to-end learning'') or improving/refining existing solutions (similar to ``machine learning alongside optimisation algorithms'' discussed above). A training dataset including instances from the optimisation problems is usually used to evaluate the quality of evolved heuristics. Recent work in \cite{Nguyen21GCP} suggests a simple way to use GP to enhance the efficiency of constraint programming by designing effective variable ordering heuristics. A number of advanced GP algorithms have been proposed to improve the quality of heuristics used to solve combinatorial optimisation problems \cite{Zhangsurevy2023}. Recent studies on GP with machine learning techniques often examine feature selection, surrogate modelling, ensemble learning, and multitask learning \cite{ZhangMT23}.

\subsection{Genetic programming algorithms for RCJS}
Recently, GP has also been applied to solve different variants of RCJS. \cite{CHAND2018146} apply GP to evolve priority dispatching rules for resource-constrained project scheduling and show that the evolved rules are significantly better than existing rules and heuristics. In their paper, priority rules are functions that are used to assign priorities to project activities based on their characteristics, such as durations and release times. The study by~\cite{Nguyen2018geccomultipass} is the pioneering research that applies GP to develop heuristics for the RCJS. Two representations based on simple expression trees and iterative rules \cite{nguyen_learning_2013} are adopted to generate heuristics for RCJS. The experiments show that evolved heuristics can quickly produce good solutions, especially for large instances. The main limitation of the method is that there is still a large gap between solutions generated by evolved heuristics and optimal solutions. Moreover, in these studies, a solution construction heuristic is always needed to generate complete and feasible solutions. In the method proposed in this paper, solution construction heuristic is not needed.

\section{Proposed method}
\label{sec:gcp}
\begin{figure*}[!tb]
\centering
\includegraphics[clip,width=0.8\textwidth]{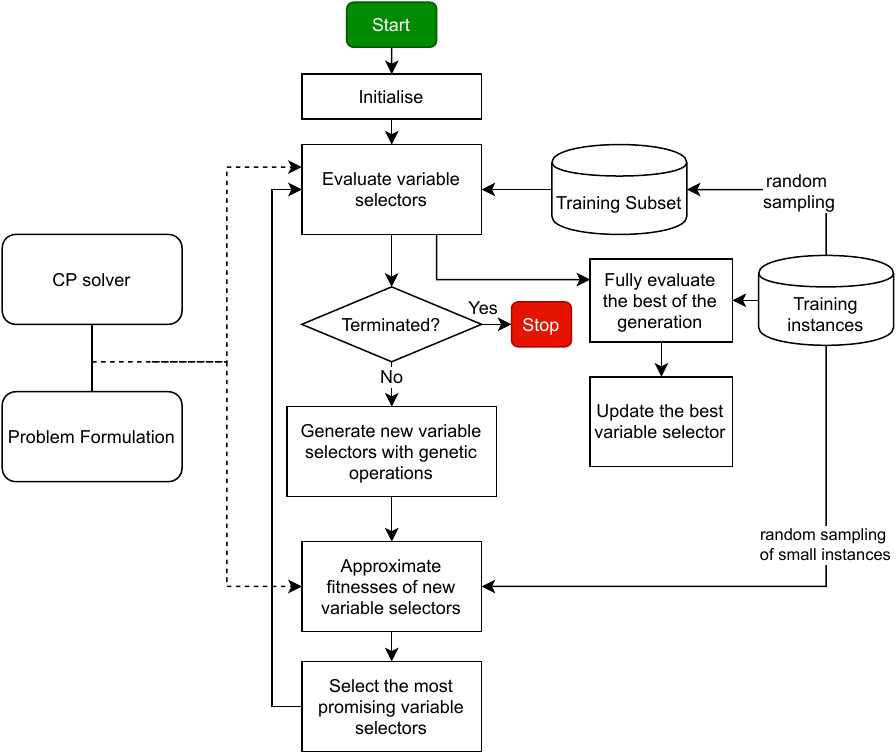}
\caption{An overview of our proposed GCP algorithm.}
\label{fig:gcp}
\end{figure*}

Figure~\ref{fig:gcp} shows a high-level overview of our proposed genetic-based constraint programming (GCP) algorithm. The algorithm starts by generating individuals, which are variable selectors, in a random manner. The population of variable selectors, denoted as $\mathbb{S} = \{\mathcal{S}_1, \dots, \mathcal{S}_{PS}\}$, where $PS$ represents the population size, is then evaluated using a CP solver along with the given problem formulation. In this study, we use Google OR-Tools\footnote{\url{https://developers.google.com/optimization}} as the CP solver. The fitness of variable selectors is determined by the quality of solutions (i.e., $TWT$) obtained by CP for a subset of training instances $\mathcal{I}_g$, within a fixed time limit. The selector $\mathcal{S}^*_g$ with the best fitness in the current generation is then evaluated using the complete training dataset $\mathcal{I}$ to avoid overfitting. If $\mathcal{S}^*_g$ shows improvement, it replaces the best-so-far selector $\mathcal{S}^*$. Individuals in the population are selected for crossover and mutation based on their fitness evaluations to generate new offspring. A simple pre-selection technique is employed to prevent the generation of poor-performing variable selectors in the next generation. This technique evaluates variable selectors using small instances (small numbers of jobs and machines), selecting only the top $PS$ selectors from these trials to be moved to the next generation. The algorithm terminates when the maximum generation is reached. The details for each key component of our GCP algorithm will be provided in the following subsections.


\subsection{Representation}
\label{sec:representation}

\begin{figure*}[!tb]
\centering
\includegraphics[clip,width=0.88\textwidth]{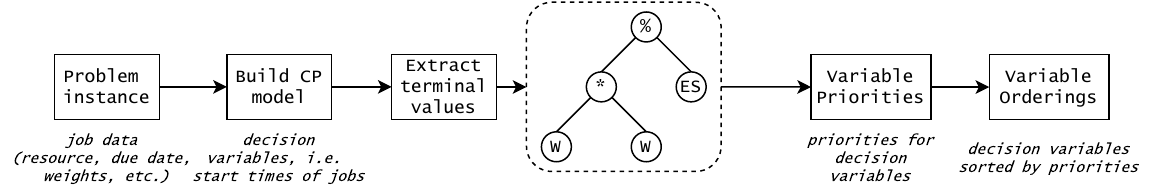}
\caption{Representation of variable selector.}
\label{fig:vsrepresentation}
\end{figure*}

We define the CP model $\Omega = (\mathcal{X}, \mathcal{D}, \mathcal{C})$ to solve an instance of the RCJS problem. All variables in the problem are represented by $\mathcal{X} = \{x_1, \dots, x_n\}$, that is, the start times of all jobs. The domains of the variables are represented by $\mathcal{D} = \{\mathcal{D}_1, \dots, \mathcal{D}_n\}$,  where $x_k \in \mathcal{D}_k$ or the start time variables have discrete domain values between 0 and $\mathcal{T}$. For example, for job $k$, $x_k \in [0,\ldots,\mathcal{T}-p_j]$. The constraints are represented by the set $\mathcal{C} = \{C_1, \dots, C_L\}$, which includes precedence or disjunctive constraints. Each constraint $C_l$ is defined by a relation $\mathcal{R}_l$ and a subset of variables $\mathcal{X}_l \in \mathcal{X}$, which determine the values to which the variables $x \in \mathcal{X}_l$ can be assigned. To find a solution to $\Omega$, the CP solver has a search strategy that decides: (1) which variable to branch on next, and (2) an appropriate domain value to assign to the chosen variable. The GCP component attempts to identify which variable to select next. For the purposes of this study, there are two sub-strategies for variable selection, which are {\it variable ordering} and {\it domain-based variable selection}.

First, in domain-based variable selection, the domain values of all variables are considered to select which one to branch on next. As a part of the search method of CP solvers, there will typically be a default strategy, but also allow different options for domain-based variable selection. For example, those variables with the smallest domain size or minimum domain value will be selected next as part of the CP solver's default strategy. Variable ordering, on the other hand, is applied as a tie-breaker if the domain-based variable selection cannot lead to a unique variable to branch on. The motivation behind these sub-strategies is aimed at identifying feasibility (picking variables with a few remaining domain values, thereby satisfying constraints) or identifying high-quality solution (by carefully considering the order of variables). Our proposed GCP approach focusses on variable ordering with the aim of guiding CP towards finding higher-quality solutions. In the GCP-evolved variable selectors, the fixed domain-based variable selection is obtained via \textit{choosing-lowest-domain-value}, while the variable ordering is evolved. In order to reduce the domains of variables (domain reduction strategy), the CP model uses \textit{selecting-min-value}.

Variable ordering is performed by calculating the priorities for each decision variable with a priority function. In GCP, priority functions are expression trees in which terminal nodes are characteristics of jobs in the RCJS directly extracted from the CP model, and non-terminal nodes are arithmetic or logical operators. Figure~\ref{fig:vsrepresentation} shows an example of a priority function. The list of terminals and functions used to compose priority functions is shown in Table~\ref{tab:tfset}. Most termimal values, such as $ES, PT, W, DD$, can be obtained directly from problem instances while values of high-level terminals such as $WL, maxWL, NPREC, NSUC, WLPREC, WLSUC$ can be obtained when CP models are built. For the function set, GCP uses a set of arithmetic and logical functions that are commonly used in the literature \cite{branke_automated_2016}. This representation allows GCP to explore a wide range of priority functions and determine those that produce efficient variable orderings for CP. Unlike \cite{Nguyen2018geccomultipass} in which priority functions are applied multiple times to construct complete solutions for each instance, the priorities of variables are only calculated once in GCP and used as input for the CP solver.

\begin{table}
\caption{The terminal set and function set (corresponding to a job $j$) used in GCP.}
\label{tab:tfset}
\centering
\begin{tabular}{ll}  
\toprule
Notation    & Description \\
\midrule
$ES$      & earliest start time or release time     \\
$PT$      & processing time     \\
$W$      & weight    \\
$DD$      & due date  \\ 
$WL$      & workload of the machine processing $j$     \\ 
$maxWL$      & maximum workload for all machines      \\
$NPREC$      & number of precedence jobs      \\
$NSUC$      & number of successors      \\
$WLPREC$      & workload of precedence jobs      \\
$WLSUC$      & workload of successors      \\
 \midrule
$Functions$      & $+, - , * , \%, \max, \min$          \\
\bottomrule
\end{tabular}
\end{table}

\subsection{Fitness evaluation}
In \cite{Nguyen21GCP}, the fitness of an evolved variable selector $\mathcal{S}$ is based on computational efforts, that is, the number of branches that the CP solver uses to optimally solve a set of training instances. If a variable selector uses a smaller number of branching steps to reach optimal solutions, that selector will have a better fitness. However, this fitness evaluation method is only feasible if the training instances are small or solvable by the CP solver. Another issue with this method is overfitting as shown in \cite{Nguyen21GCP}. To overcome these issues, we propose here a new fitness function based on solution quality rather than computational efforts. The fitness of a variable selector $\mathcal{S}$ can be calculated as follows:
\begin{equation}
    fitness(\mathcal{S}, \mathcal{I}_g) = \frac{1}{|\mathcal{I}_g|}\sum_{I \in \mathcal{I}_g}{obj(\mathcal{S}, I, T)},
    \label{eq:fitness}
\end{equation}
\noindent where $\mathcal{I}_g$ is the set of instances randomly sampled from the training dataset $\mathcal{I}$ for fitness evaluations in generation $g$, and $obj(\mathcal{S}, I, T)$ is the objective value for $I$ obtained by the CP solver after a solving time limit $T$. We prioritise evolved variable selectors that assist the CP solver in rapidly identifying high-quality solutions within limited time frames rather than focusing on achieving and proving optimal solutions. This approach is appropriate for scenarios where problems, such as the RCJS, are difficult to solve optimally and require the generation of feasible, high-quality, and quick solutions. Furthermore, this method ensures that the GCP training times are more predictable than the approach used in \cite{Nguyen21GCP}.


To further improve the training efficiency of GCP, we adopt the sampling strategy used in \cite{hildebrandt_towards_2010,Nguyen_PES_2019} for fitness evaluations. For each generation, only a subset $\mathcal{I}_g$ of the complete training set $\mathcal{I}$ is randomly sampled for fitness evaluations. This strategy has been shown to improve the population diversity of GP algorithms. In a generation $g$, the evolved selector $\mathcal{S}_g^*$ with the best $fitness(\cdot, \mathcal{I}_g)$ will be fully evaluated with $\mathcal{I}$ and will replace the best-so-far selector $\mathcal{S}^*$ if an improvement is made.

\subsection{Genetic operators}
Traditional subtree crossover and subtree mutation~\cite{poli_field_2008} are used in our proposed GCP algorithm to produce offspring or new variable selectors. Parents for these genetic operators are selected using a tournament selection. For crossover, a subtree is randomly selected from each parent, and two subtrees are swapped to produce offspring. For mutation, a random subtree is selected from the selected parent and will be replaced with an entirely new random subtree. 

\subsection{Pre-selection technique}
Since the genetic operators used in GP are highly random, there is a high chance that poorly performing individuals will be generated. To avoid spending computational budgets evaluating bad individuals, pre-selection techniques have been applied as a screening step in which only the most promising individuals are kept for the next generation. In our study, we developed a new pre-selection algorithm in Algorithm~\ref{alg:preselect}. An intermediate population $\mathbb{S}'$ with newly generated individuals based on the above genetic operators will go through a trial process, in which small instances are used to quickly determine their quality. Another advantage of this pre-selection process is to evaluate the ability to reach optimal or near-optimal solutions of evolved variable selectors, which is not a main emphasis in the proposed fitness function.

\begin{algorithm}[tb]
\caption{Pre-selection algorithm}
\label{alg:preselect}
\def\NoNumber#1{{\def\alglinenumber##1{}\State #1}\addtocounter{ALG@line}{-1}}
\begin{algorithmic}[1]
\Statex \textbf{Input:} A set of newly generated variable selectors $\mathbb{S}'$ ($|\mathbb{S}'| > |\mathbb{S}|$), and training dataset $\mathcal{I}$
\Statex \textbf{Output:} a new population $\mathbb{S}$
\State $\mathcal{I}' \gets$ select small instances from $\mathcal{I}$
\State $\mathbb{S} \gets \emptyset$
\For {$\mathcal{S} \in \mathbb{S}'$}
    \State $fitness(\mathcal{S}, \mathcal{I}') \gets$ evaluate $\mathcal{S}$ with new instances in $\mathcal{I}'$
\EndFor
\State sort $\mathbb{S}'$ based on $fitness(\cdot, \mathcal{I}')$ (lowest to highest)
\State $\mathbb{S} \gets$ select the top $PS$ selectors from $\mathbb{S}'$
\State return the new population $\mathbb{S}$
\end{algorithmic}
\end{algorithm}

\section{Experimental Design}
\label{sec:exp}

\subsection{Datasets}
For the experimental evaluation, we make use of two datasets. The first one consists of 1755 randomly generated instances obtained by the instance generator of \cite{nguyen_cor_2019}.\footnote{\url{https://github.com/andreas-ernst/Mathprog-ORlib/blob/master/data/RCJS_new_instances.zip}} The dataset includes combinations of different characteristics of RCJS with the following settings: 
\begin{itemize}
\item Number of machines:  \{3, 4, 5, 6, 7, 8, 9, 10, 12, 15, 18, 20, 25\},  
\item Network complexities (precedence probability): \{0.1, 0.2, 0.3, 0.4, 0.5, 0.6, 0.7 0.8, 0.9\},
\item Resource utilisation: \{0.25, 0.5, 0.75\}. 
\end{itemize}
Using the study by \cite{singhernst10} as a guide, each machine consists of approximately 10.5 jobs. This dataset was used for training and testing purposes. The second dataset that we investigate is a well-known benchmark \cite{singhernst10,ernstsingh12,Thiruvady2014,Thiruvady2016}, which has provided the baseline for comparing solution methods in the literature.\footnote{\url{https://github.com/andreas-ernst/Mathprog-ORlib/blob/master/data/RCJS_Instances.zip}} In the following sections, we refer to the first dataset as the \textit{nd-rcjs} dataset and the second one as the \textit{se-rcjs} dataset (the datasets are named after the authors' initials).

\subsection{Parameter settings}
Table~\ref{tab:params} shows the parameters of GCP and the CP solver. These parameters are similar to those used in \cite{Nguyen21GCP} and are further fine-tuned through pilot experiments. We then set the number of generations to 50, which is sufficient for the population to converge. In our experiments, to avoid biases caused by instances with different sizes, for our training set, we fix the number of machines in the chosen instances. Specifically, we only use instances with 6 machines for fitness evaluations. Instances with 6 or fewer machines will be used for the pre-selection step. These instances are selected based on our observations in previous studies that RCJS instances with 6 machines or more are unlikely to be solved to optimality.

\begin{table}
\caption{Parameter settings}
\label{tab:params}
\centering
\setlength{\tabcolsep}{4pt}
\begin{tabular}{llr}  
\toprule
& Parameter   & Value \\ \midrule
GCP & Population size & 200, 500, 1000 \\
& Number of generations & 50 \\
& Terminal and function sets & see Table~\ref{tab:tfset}\\
& Initialisation & ramped-half-and-half \\
& Tournament selection size & 5 \\
& Subtree crossover & 90\% \\
& Subtree mutation & 10\% \\
& Maximum depth & 7 \\
& Intermediate population size & $= 2 \times $ population size \\
& \# of training instances $|\mathcal{I}|$ & 81 \\
& \# of instances per generation $|\mathcal{I}_g|$ & 5 \\
& \# of instances for pre-selection & 10 \\ \midrule
CP & Solver & Google Or-tools \\
 solver & Solving time limit (training) & 30 (s) \\
& Solving time limit (testing) & 60 (s) \\
\bottomrule
\end{tabular}
\end{table}

When solving each instance in the training or testing step, the CP solver terminates when the time limit expires, and the solution with the best objective value at that point will be reported. Feasibility is not an issue, since for all RCJS instances we use, the CP solver can always find at least one feasible solution in the allowed time-frame. The time limit for testing purposes is allowed to be a bit larger in order to examine whether the advantages of evolved variable selectors can be realised in the later stages of the CP search process. The default domain-based variable selection and the default domain reduction are \textit{choosing-lowest-domain-value} and \textit{selecting-min-value}, respectively. The variable selectors evolved by GCP will be used as tie-breakers to help \textit{choosing-lowest-domain-value} decide the next variable to branch. 

\subsection{Compared algorithms}
To validate the effectiveness of the proposed algorithm, we compare GCP with other GP and optimisation methods. Other GP algorithms compared in our experiments include GP for evolving a single-pass heuristic and GP for iterative dispatching rules (IGP) \cite{Nguyen2018geccomultipass}. Compared to the single-pass heuristic, we examine the performance of evolved selectors against traditional construction heuristics. On the other hand, IGP aims at evolving improvement heuristics, which will be a good candidate for verifying the benefits of combining GP and CP. We also attempt to compare GCP with the GP algorithm proposed in \cite{Nguyen21GCP}. However, RCJS instances are too complex to solve to optimality; therefore, the GP algorithm proposed in \cite{Nguyen21GCP} failed to make any significant progress as its fitness function cannot differentiate between good and bad variable selectors. Optimisation algorithms compared in our experiments are ant colony optimisation (ACO) \cite{Thiruvady2014}, and column generation and ACO (CGACO) \cite{Thiruvady2016}. These two algorithms specialise in solving RCJS instances and employ a wide range of optimisation tricks to improve the search efficiency. Comparing our algorithms to these two optimisation algorithms helps access the value of learning efficient variable selectors.

\section{Results}\label{sec:results}
This section compares the results of GCP with existing GP algorithms and specialised optimisation methods that have been proposed for RCJS in the literature. Since RCJS instances are well-known for their complexity, our paper focuses on comparing the solution quality (objective values of best-found solutions) rather than the number of optimal solutions. The impact of population size on the effectiveness of evolved variable selectors is also examined. The population size represents the computational effort required to evolve variable selectors. However, as training or evolutionary processes can be done offline, large population sizes will not influence the solving time of the CP solver, which only uses the evolved variable selector. In addition, computational times to execute evolved variable selectors when solving each instance are very small (less than a fraction of a second) and will not be discussed.

\begin{figure}[!tb]
\centering
\subfigure[Objective values]{\includegraphics[width=0.4\textwidth]{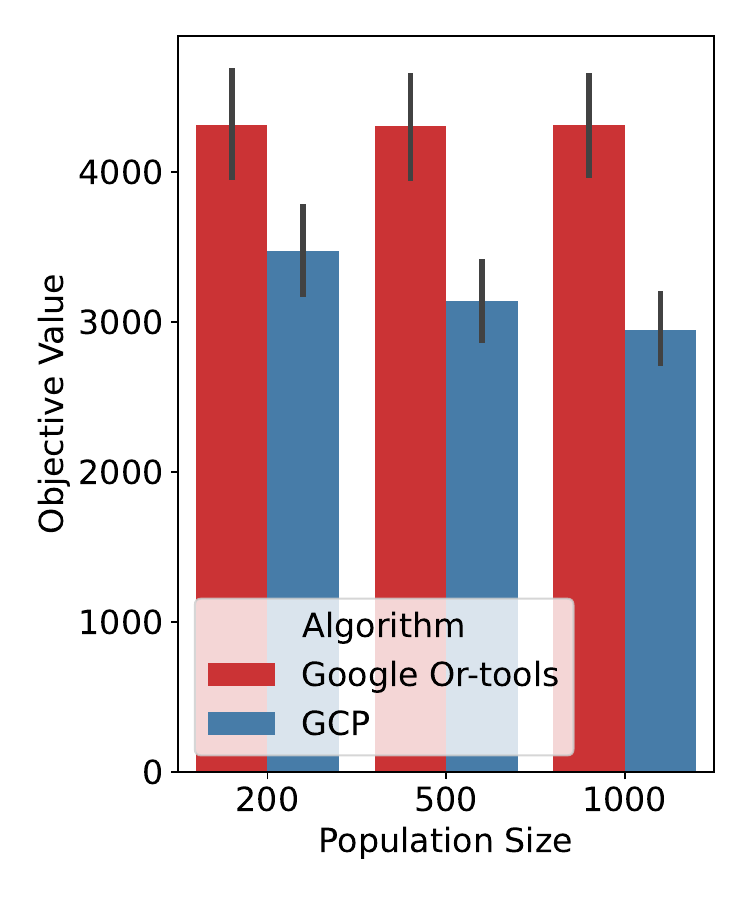}}
\subfigure[\% Optimal solutions]{\includegraphics[width=0.4\textwidth]{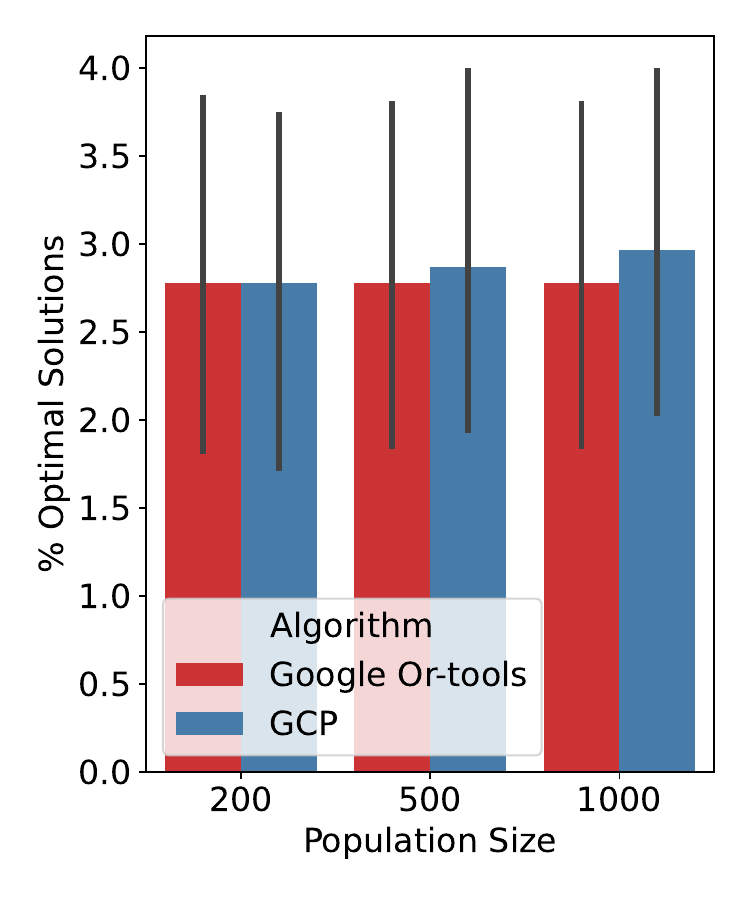}}
\caption{The comparison between our proposed GCP and the default setting of Google OR-Tools (\textit{se-rcjs} dataset).}
\label{fig:defaultcpcompare}
\end{figure}

\subsection{Comparisons to the default CP settings}
Figure~\ref{fig:defaultcpcompare} shows the average objective values and percentage of optimal solutions obtained by Google's OR-Tools CP solver and GCP with different population sizes. Note that the instances used in these experiments (from \textit{se-rcjs} dataset) are not seen during the training process of GCP. The results suggest that GCP can produce better solutions than its counterpart and slightly improves the chances of proving optimality. These results demonstrate both the effectiveness of GCP and the generalisation of its evolved variable selectors. It is worth noting that increasing computation effort can greatly benefit GCP. From Figure~\ref{fig:defaultcpcompare}, it is clear that large population sizes can lead to more effective variable selectors.

The advantages of GCP are different across different subsets, as seen in Figure~\ref{fig:gpcompare} and Figure~\ref{fig:optcomparerandom}. For small instances with fewer than 7 machines, GCP does not show any clear advantage as the CP solver can determine optimal or near-optimal solutions relatively easily regardless of variable orderings. As the size of the problem increases, we can see the importance of variable orderings. For the largest instances in the \textit{se-rcjs} dataset, the objective values obtained by GCP are nearly half of those obtained by Google OR-Tools. These results confirm that GCP can enhance the efficiency of the CP solver, and moreover, increased improvements are seen with increasing problem sizes.   

\subsection{Comparisons to GP algorithms}
\begin{figure}
\centering
\subfigure[Instances with less than 9 machines]{\includegraphics[width=0.48\textwidth]{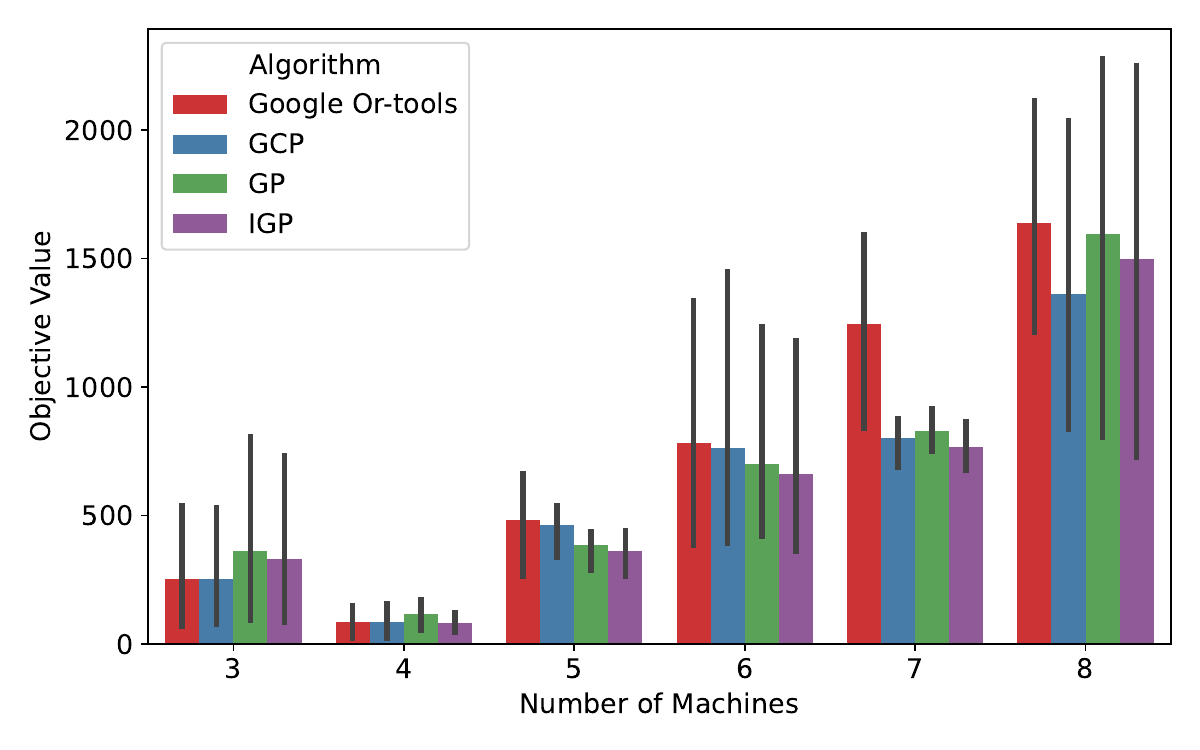}}
\subfigure[Instances with 9 machines or more]{\includegraphics[width=0.48\textwidth]{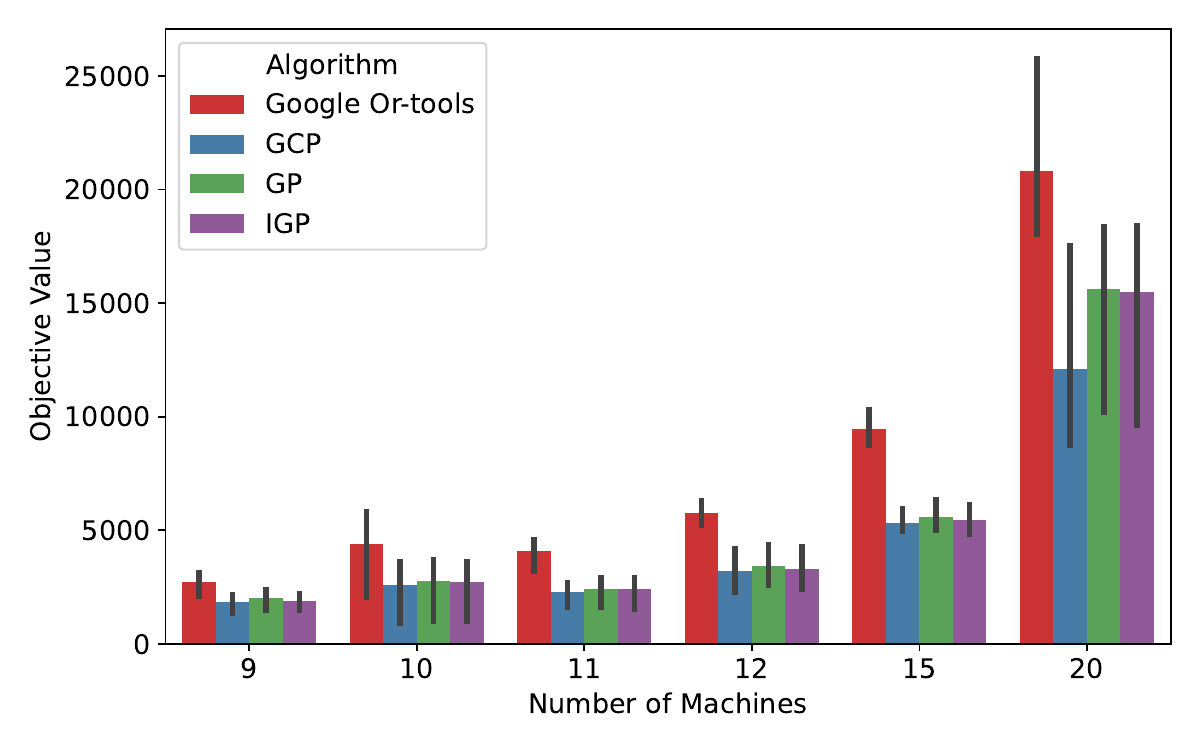}}
\caption{The comparison between GCP and GP algorithms.}
\label{fig:gpcompare}
\end{figure}

Figure~\ref{fig:gpcompare} shows the performance of GCP compared to simple GP (or GP for short) and IGP, proposed in \cite{Nguyen2018geccomultipass}. The other two GP algorithms in previous studies are only trained and tested with the {se-rcjs} dataset, so these algorithms are expected to be more specialised for the instances in our comparison. The results show no clear difference between GCP and other GP algorithms when dealing with small and medium instances (number of machines fewer than 12) -- except for the instances with 3 machines. For larger instances (with 12 machines or more), GCP shows a clearer advantage. For large instances with 20 machines (the hardest instances in the {se-rcjs} dataset), the solutions obtained by GCP are 20\% lower than those of GP and IGP. 

Given that the terminal sets used by GCP and the other two algorithms are equivalent, the improvements seen here are directly attributable to the combination of GP and CP. For small instances, the CP solver can complete an exhaustive search to determine optimal solutions, which are very difficult to obtain with the heuristics evolved by the other two GP algorithms. For large instances, the CP solver again gives GCP a better chance to find improvements, via the systematic search of CP, rather than the construction and improvement procedures employed in the other two GP algorithms.

Apart from the performance advantages, GCP works directly with user-friendly problem formulations and does not require low-level heuristics such as solution construction and improvement heuristics, which are necessary for the other GP algorithms. This has several practical implications. First, users can easily address new situations (e.g. additional technological constraints) by modifying the problem formulation rather than developing new customised heuristics, which can be time-consuming (for testing and fine-tuning). Second, if larger run-times are allowed, GCP can make further attempts to identify better solutions, which is impossible with heuristics evolved by the other GP algorithms.

\subsection{Comparisons to optimisation algorithms}
\begin{figure}
\centering
\subfigure[Instances with less than 9 machines]{\includegraphics[width=0.48\textwidth]{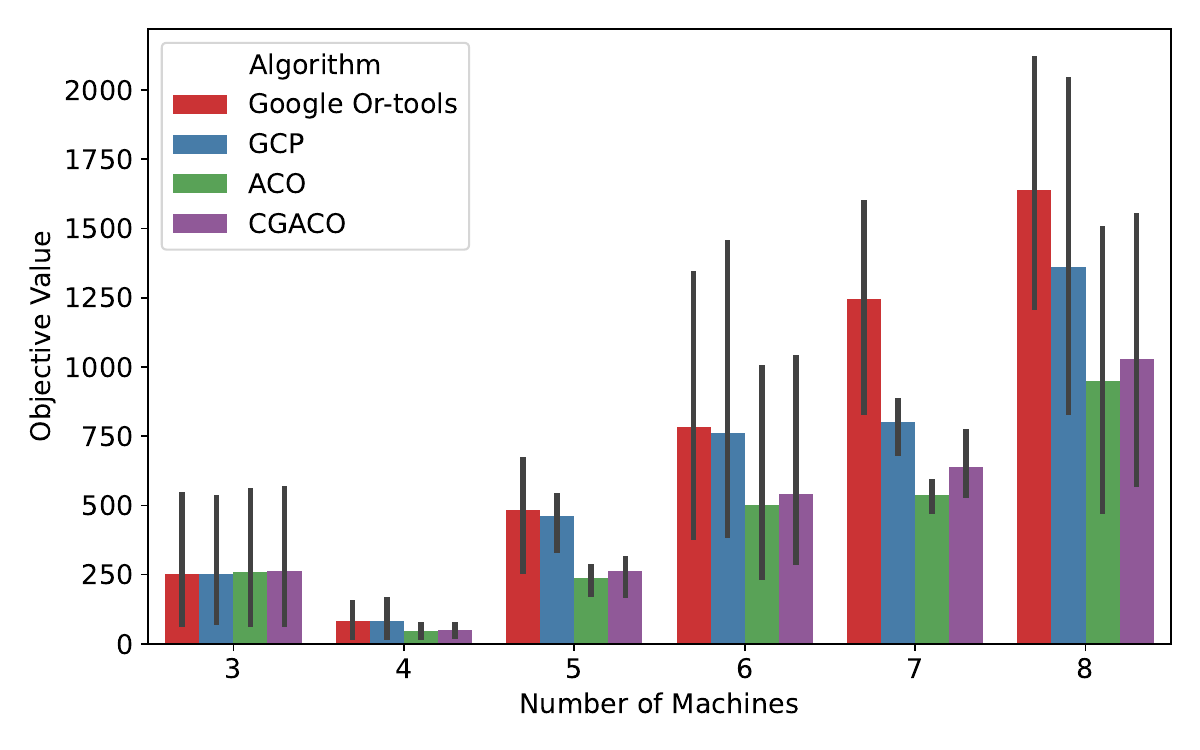}}
\subfigure[Instances with 9 machines or more]{\includegraphics[width=0.48\textwidth]{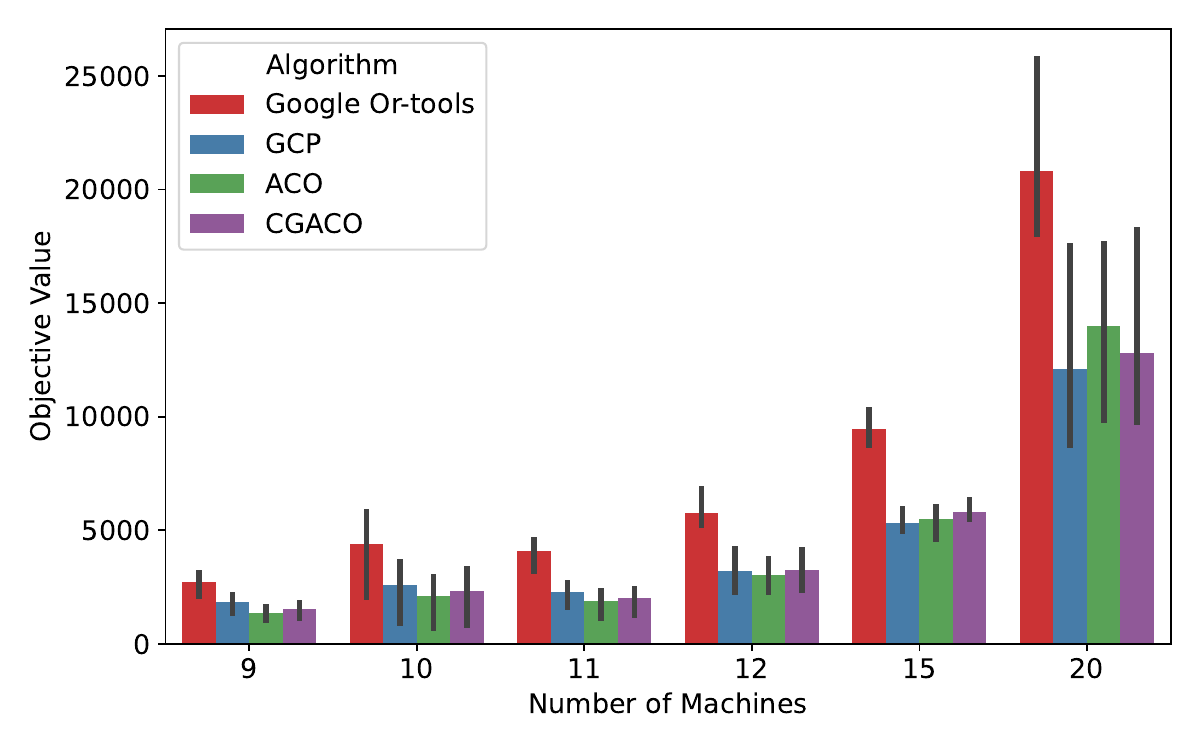}}
\caption{A comparison of GCP and other specialised optimisation algorithms.}
\label{fig:optcomparerandom}
\end{figure}

In Figure~\ref{fig:optcomparerandom}, the evolved variable selectors are compared to other specialised optimisation algorithms for RCJS. It is noted that the results reported for ACO and CGACO are obtained after running these algorithms for 60 minutes, which is 60 times more expensive than GCP (only 60 seconds during the testing phase). For the \textit{se-rcjs} dataset, ACO and CGACO are better than GCP for the small and medium instances (fewer than 12 machines). However, GCP becomes more competitive for instances with 15 machines and outperforms ACO and CGACO for instances with 20 machines. Achieving these results with a much smaller running times show that GCP is a powerful tool to solve RCJS problem.

Given that ACO and CGACO \cite{nguyen_cor_2019} are very sophisticated algorithms with many specialised heuristics designed for RCJS and the support of advanced integer programming solvers, these results are very encouraging. One key advantage of GCP is that it incorporates little or no problem domain knowledge manually into the solving process, which is the key factor in the success of CGACO. Therefore, GCP can be easily generalised to work with different variants of RCJS without major modifications.  


\subsection{Further analyses}
\begin{figure}
\centering
\includegraphics[width=0.7\textwidth]{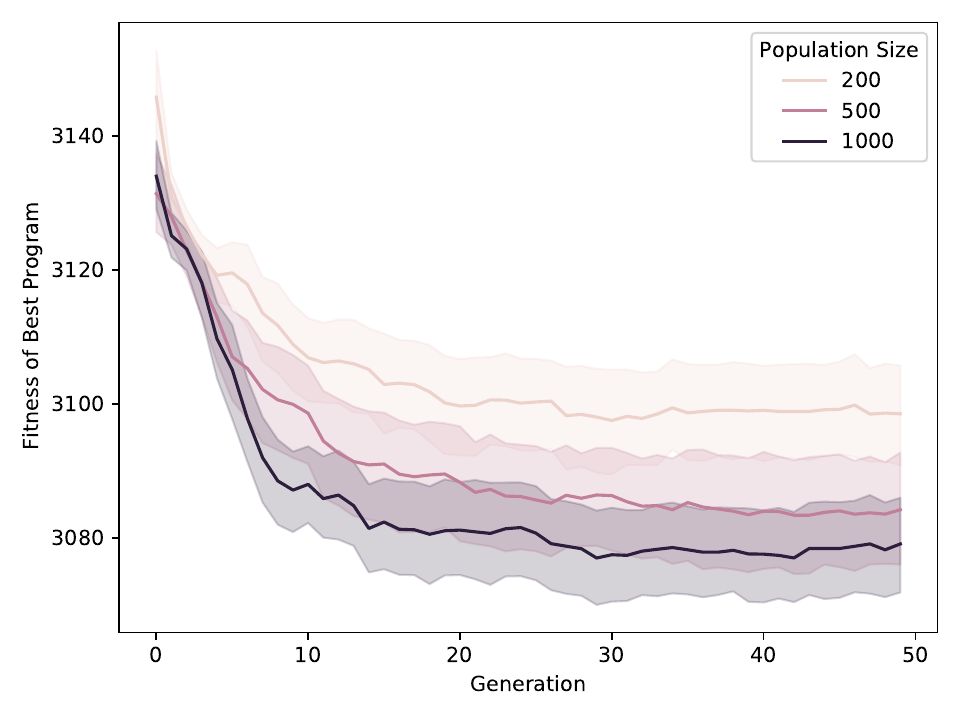}
\caption{The training progress of GCP.}
\label{fig:traingbest}
\end{figure}

Figure~\ref{fig:traingbest} sheds some light on the training or evolutionary process of GCP. GCP with three different population sizes can converge within 50 generations, and the results show that a larger population helps GCP identify better variable selectors (low fitness values). These observations are consistent with the testing results shown in Figure~\ref{fig:defaultcpcompare} and indicates  no serious overfitting issue in GCP.

\section{Conclusions and future studies}
\label{sec:conclude}

In this study, we propose an automated heuristic design method based on GP to enhance the performance of constraint programming solvers to tackle resource-constrained job scheduling. The idea underlying our method is to use GP to search over a heuristic space, thereby learning variable orderings that can be used within the search component of CP solvers. An experimental evaluation on two benchmark datasets shows the following results: (a) CP with the evolved variable selectors (GCP) is more effective than Google's OR Tools SAT solver in terms of solution quality and proving optimality, (b) GCP is more effective than existing GP techniques for RCJS,  and (c) GCP is competitive with existing optimisation methods specifically designed for RCJS and able to outperform these methods for large instances with a much smaller running time.

Our proposed GCP approach has demonstrated promise on RCJS and has previously shown to be effective on job shop scheduling. It could prove to be effective on RCJS with uncertainty~\cite{THIRUVADY2022sacs, thiruvady2022adaptive}, which the authors plan to investigate as part of future work. Moreover, being problem independent, we conjecture that this approach can be readily transferred to any problem that can be modelled efficiently as a constraint program. In doing so, we can expect to find higher quality solutions in reduced time-frames. 

In the context of RCJS, we have seen the excellent performance of our proposed method. However, it is possible to further improve the method, for example, by considering other machine learning approaches or making use of other problem specific heuristics in the terminal set. A more ambitious direction is to develop automated methods to directly extract high-level patterns, such as separability, from CP models, which can lead to huge efficiency improvements.

\bibliographystyle{plain}
\bibliography{sample-base,gcp}

\end{document}